\documentclass{article}

\usepackage{microtype}
\usepackage{graphicx}
\usepackage{subfigure}
\usepackage{booktabs} 
\usepackage{amssymb}
\usepackage{amsmath,tabu,amsthm}

\usepackage{hyperref}

\usepackage[accepted]{style}

\newcommand{\T}{\mathcal{T}}
\newcommand{\A}{\mathcal{A}}
\newcommand{\F}{\mathcal{F}}
\newcommand{\res}{\text{res}}
\newcommand{\incon}{\text{Incon}}

\theoremstyle{definition}
\newtheorem{definition}{Definition}[section]

\icmltitlerunning{A Topological-Framework to Improve Analysis of Machine Learning Model Performance}

\begin{document}

\twocolumn[
\icmltitle{Dataset to Dataspace: A Topological-Framework to Improve Analysis of \\Machine Learning Model Performance}



\icmlsetsymbol{equal}{*}

\begin{icmlauthorlist}
\icmlauthor{Henry Kvinge}{PNNLSea,UW}
\icmlauthor{Colby Wight}{PNNLRich}
\icmlauthor{Sarah Akers}{PNNLRich}
\icmlauthor{Scott Howland}{PNNLRich}
\icmlauthor{Woongjo Choi}{PNNLRich}
\icmlauthor{Xiaolong Ma}{PNNLRich}
\icmlauthor{Luke Gosink}{PNNLRich}
\icmlauthor{Elizabeth Jurrus}{PNNLSea}
\icmlauthor{Keerti Kappagantula}{PNNLRich}
\icmlauthor{Tegan H. Emerson}{PNNLSea,CSU}
\end{icmlauthorlist}

\icmlaffiliation{PNNLSea}{Pacific Northwest National Laboratory, Seattle, WA, USA}
\icmlaffiliation{UW}{Department of Mathematics, University of Washington, Seattle, WA, USA}
\icmlaffiliation{CSU}{Department of Mathematics, Colorado State University, Fort Collins, CO, USA}
\icmlaffiliation{PNNLRich}{Pacific Northwest National Laboratory, Richland, WA, USA}

\icmlcorrespondingauthor{Henry Kvinge}{henry.kvinge@pnnl.gov}

\icmlkeywords{Model robustness, topology, presheaves, model evaluation}

\vskip 0.3in
]



\printAffiliationsAndNotice{}  

\begin{abstract}
As both machine learning models and the datasets on which they are evaluated have grown in size and complexity, the practice of using a few summary statistics to understand model performance has become increasingly problematic. This is particularly true in real-world scenarios where understanding model failure on certain subpopulations of the data is of critical importance. In this paper we propose a topological framework for evaluating machine learning models in which a dataset is treated as a ``space'' on which a model operates. This provides us with a principled way to organize information about model performance at both the global level (over the entire test set) and also the local level (on specific subpopulations). Finally, we describe a topological data structure, {\emph{presheaves}}, which offer a convenient way to store and analyze model performance between different subpopulations.
\end{abstract}

\section{Introduction}
\label{sect-introduction}

Advances in deep learning have resulted in major breakthroughs in a range of machine learning tasks \cite{lecun2015deep}. The price of such advances have been dramatic increases in model size and complexity, fueled by ever larger training and test sets. As a consequence, it has become more challenging to understand model performance. In the literature, models are frequently judged based on a few statistics that are calculated over an entire test set. The use of such an evaluation scheme can lead to lack of robustness through underspecification \cite{d2020underspecification}. It can also obscure serious model failures on certain subpopulations of the data \cite{buolamwini2018gender,yao2011combining,recht2019imagenet}. 

In this paper we propose one approach to begin to address this issue, based on the observation that a ``dataset'' is rarely just a set of points. Instead, most datasets have a significant amount of metadata attached to them. In the supervised setting, this could include labels, but it might also include a wide variety of additional information. For example, from an image classification dataset we might have access to the location and time an image was taken, the kind of camera that was used or resolution of the image, or other objects or properties labeled in the image. It was observed in \cite{kvinge2021sheaves} that the presence of such metadata means that we should treat a dataset as a ``space'' rather than just a set, with proximity between datapoints based on similarities or differences in labels and metadata. In this paper we show how this paradigm can provide a framework for principled, fine-grained analysis of model performance.

Following \cite{kvinge2021sheaves}, we begin by recalling how a topology, which encodes the bare essentials necessary to define a notion of space, can be built on top of a dataset. We remind the reader of {\emph{presheaves}}, a data structure from topology that allows one to attach data locally (that is, to subpopulations of a dataset). We show how these constructions can be used to (i) systematically keep track of model performance across many subsets of related points and (ii) compare model performance on non-disjoint subsets. We finally show that presheaves allow us to make sense of a range of questions about a dataset which would otherwise be ill-defined. For example: {\emph{``How does a model perform in the vicinity of a datapoint $x$?''}} We conclude with an exploration of these ideas using a ResNet18 convolutional neural network \cite{he2016deep} trained on the Caltech-UCSD Birds $200$ dataset \cite{WahCUB_200_2011}. We use the extensive image attribute metadata associated with this dataset to build a topology which reflects similarities in bird appearance that go beyond information contained in image labels alone. While many of the analytical approaches presented in this paper are actually rather simple and could have been developed without the underlying topological machinery, we believe that building a precise mathematical framework in which to work makes it easier to conceive of and develop new approaches to evaluating model robustness across large and complex datasets.

\section{Related work}
\label{sect-related-work}

Model robustness has become an increasingly important topic as deep learning models have begun to be deployed in a range of safety-critical applications. Some of this work has focused on the extent to which models are robust to perturbation or corruption of input \cite{hendrycks2019benchmarking} or shifts in distribution \cite{hendrycks2020many}. The present work is inspired by a line of research investigating how models can lack robustness when they systematically fail on certain subpopulations of a dataset \cite{oakden2020hidden}. There are a range of approaches to mitigating this phenomenon, including the use of novel loss functions or methods of training that promote more robust performance on specified or unspecificed subpopulations \cite{duchi2020distributionally,sohoni2020no}. Our work can be seen as a complementary approach to these methods, creating a framework in which one can systematically organize subpopulations of a dataset for the purposes of analyzing and mitigating undesirable model behavior.

In the last $10$ years there has been a push to find ways of applying tools from the field of topology to questions in data science \cite{carlsson2009topology}. Much of the resulting work, commonly known as {\emph{topological data analysis (TDA)}}, has focused on developing methods of measuring the ``shape'' of point clouds via notions from topology such as homology \cite{edelsbrunner2000tds,zomorodian2005computing}. This is distinct from the present work which takes a more combinatorial approach to topology, choosing to build a finite topology \cite{barmak2011algebraic} induced by metadata.

Presheaves have only recently begun to be applied to problems in machine learning and data science. They have, for example, been used for uncertainty quantification in geolocation \cite{joslyn2019sheaf}, air traffic control monitoring \cite{mansourbeigi2017sheaf}, learning signals on graphs \cite{hansen2019learning}, and data fusion \cite{robinson2017sheaves}. This is the first time that this data structure has been applied to the problem analyzing model performance.

\section{Background}
\label{sect-background}
\subsection{Topologies and presheaves}

This paper will utilize two foundational constructions from mathematics (1) the notion of a topology and (2) the notion of a presheaf. Due to length limitations, in this paper we confine ourselves to a concise definition of each, noting that the former is an entire discipline in mathematics and the latter is a ubiquitous tool that appears across many fields of mathematics. We urge the interested reader to consult \cite{munkres2014topology,hatcher2002algebraic} for further information about topology and \cite{vakil2017rising} Part 1, Chapter 2 and \cite{bredon1997sheaves} for further information about presheaves.

As a consequence of the way we will use topology in this paper and the fact that all objects we deal with are finite, we give a non-standard definition of a topology based on the concept of a subbasis. 

\begin{definition}
Let $X$ be a finite set and let $B = \{U_i\}_{i \in I}$ be a collection of subsets of $X$ indexed by some set $I$. Assume that the union of all subsets in $B$ is $X$. Let $\T$ be the collection of all subsets of $X$ that can be formed by some sequence of unions and intersections of elements of $B$ (this includes the empty union). We call $\T$ the {\emph{topology}} induced by {\emph{subbasis}} $B$.
\end{definition}

The finite sets in $\T$ are called {\emph{open sets}} and are the finite analogue of open sets from more familiar topological spaces (e.g. $\mathbb{R}$). Note that it follows by construction that $\T$ is (i) closed under unions, (ii) closed under intersections, and (iii) contains $\emptyset$ (the empty union) and $X$ (the union of all elements in $B$). These conditions happen to be the axiomatic definition of a topology of a finite set \cite{munkres2014topology}. For $x \in X$, a {\emph{neighborhood of $x$}} is any open set $U \in \T$ that contains $x$. We think of $U$ as capturing some notion of the area ``around $x$''. 

A presheaf is a structure that sits on top of a topological space and allows one to systematically (i) assign data to open sets and (ii) compare the data sitting on intersecting open sets. 

\begin{definition}
Let $X$ be a finite set with topology $\T$. Then a {\emph{presheaf}} $\F$ on $\T$ is a function that to each open set $U \in \T$ associates a set $\F(U)$ (the {\emph{space of sections over $U$}}), along with a {\emph{restriction map}} $\res^\F_{U,V}: \F(U) \rightarrow \F(V)$ for each open set $V$ such that $V \subseteq U$, subject to the following conditions.
\begin{enumerate}
\item For any $U \in \T$, the trivial restriction map $\res^\F_{U,U}: \F(U) \rightarrow \F(U)$ is the identity function from $\F(U)$ to $\F(U)$.
\item For any open sets $W,V,U \in \T$ with $W \subseteq V \subseteq U$, $\res^\F_{V,W} \circ \res^\F_{U,V} = \res^\F_{U,W}$.
\end{enumerate}
\end{definition}

Note that through its restriction maps, a presheaf shadows the fact that a topological space $\T$ can be completely defined through set inclusion maps $V \hookrightarrow U$ for each pair $V,U \in \T$ with $V \subseteq U$. Restriction maps provide a way of transferring data collected on a larger region of $X$, $U$, to a smaller region, $V$. 

For open set $U$, each element of $\F(U)$ is called a {\emph{section}}. Following \cite{robinson2017sheaves}, the choice of a section $a_U \in \F(U)$, for each $U \in \T$, $\{a_U\}_{U \in \T}$, is called an {\emph{assignment}}. 

\subsection{Datasets as Topological Spaces}
\label{sect-topology-on-dataset}

Suppose that we are handed a dataset $D$. As described in Section \ref{sect-introduction}, we can often use metadata or labels to identify multiple subsets of related points from $D$. For example, if elements of $D$ have labels from set $L$, then we can form the subsets $\{U_\ell\}_{\ell \in L}$, where $U_\ell$ contains all those elements $x \in D$ with label $\ell$. Alternatively, if there are scalar values associated with elements of $D$, then for any $a \in \mathbb{R}$ we can form the set $U_{\geq a}$ (respectively, $U_{\leq a}$) which consists of all those values $x \in D$ such the scalar value associated with $x$ is greater than (resp. less than) $a$. 

Denote a choice of such subsets from $D$ by $S = \{U\}_{i \in I}$ where $I$ is some index set. As described in \cite{kvinge2021sheaves} one can form a topology $\T_S$ on $D$ by taking $S$ as a subbasis. Then $\T_S$ encodes a notion of space on $D$ that is informed by labels and other metadata from $D$. We note that even though $\T_S$ is a finite topology, it will potentially be very large and many, if not most, open sets may be hard to interpret, arising from combinations of unions and intersections of elements from $S$. We advocate putting limits on the number of intersections and unions that are actually calculated in practice. Furthermore, depending on the application one may be more interested in intersections than unions and vise versa. In the toy experiments described in Section \ref{sect-experiments} for example, we limit ourselves to intersections that include at most two of the subbasis elements. 

\section{Encoding Model Performance as a Presheaf}

\subsection{The accuracy presheaf}
\label{sect-local-performance-of-a-model}

In this section we describe a presheaf designed to store a model's accuracy on different open sets of the dataset topology $\T_S$ outlined in Section \ref{sect-topology-on-dataset}. To this end we assume that dataset $D$ is associated with a classification task with label space $L$. We note that this section is meant to function as a template for how a range of performance statistics might be encoded as presheaves. We could have chosen to use, for example, loss, precision, recall, or F1-score with only minor modification. We end the section by describing a range of ways we can use the accuracy presheaf to analyze model performance.

We create a presheaf $\A$ on the topology $\T_S$ on dataset $D$ above by setting $\A(U) = [0,1]$ (the closed interval from $0$ to $1$) for each $U \in \T_S$ with $U \neq \emptyset$ and $\A(\emptyset) = \{0\}$. For $U,V \in \T_S$ with $V \subseteq U$, we let $\res^{\A}_{U,V}$ be the identity map if $V \neq \emptyset$ and let $\res^{\A}_{U,\emptyset}$ be the zero map otherwise. We call $\A$ the {\emph{accuracy presheaf}}. Thus assignments from $\A$ consist of numbers $0 \leq a_U \leq 1$ attached to each open set $U \in \T_S$. Let $f: X \rightarrow L$ be a model that has been trained to predict the labels on data coming from the same or similar distribution as $D$. The {\emph{accuracy assignment}} $\{a_U^f\}_{U \in \T_S}$ associated with $f$, is then defined such that $a_U^f$ is the accuracy of $f$ on subset $U$ of dataset $D$. For example, if $D$ is a dataset with images of dogs, then a particular open set $U$ might consists of all dogs that have spots and are small. The value $a_U$ is then just the model's accuracy on this subset.

Beyond examining individual accuracies coming from accuracy assignment $\{a_U^f\}_{U \in \T_S}$ (for example, on which open set does $f$ achieve its highest or lowest accuracy), one can also compare how accuracies change as we move from a superset to a subset. That is, if $V \subset U$ are both open sets in $\T_S$ then we can look at the difference: $|\res^\A_{U,V}(a_U) - a_V|$, which in this case is equal to $|a_U-a_V|$ unless $V = \emptyset$ (we leave the restriction map notation in view of this being a template for other statistics that might require non-identity restriction maps). For example, how does a model's performance change when we move from images of spotted dogs to images of spotted dogs that are small? Inspired by \cite{robinson2017sheaves}, \cite{kvinge2021sheaves} introduced the notion of the local inconsistency of an assignment in an effort to measure the extent to which an assignment changes across related open sets. We define a modified version of assignment inconsistency more appropriate for studying statistics related to machine learning models. Let $\A$ be the accuracy presheaf on topological space $\T_S$ for dataset $D$. For non-negative integer $k$ and $U \in \T$, the {\emph{local $k$-bounded inconsistency at $U \in \T_S$}} of an assignment $A = \{a_U\}_{U \in \T}$ is defined as
\begin{equation*}
\incon_k(U,A) := \max_{\substack{V \subseteq U,V \in \T \\|U\setminus V| \leq k}} |\res^\F_{U,V}(a_U)-a_V|.
\end{equation*}
By including the bounding value $k$, we avoid the situation where $U$ is a very large subset of $D$ and $V$ is a very small subset of $D$. Large changes in model performance are more likely in such situations and may reflect statistical irregularities rather than model failure. 

Note that by using the notion of proximity induced by $T_S$, for any element $x \in D$, we can ask how $f$ performs in different neighborhoods of $x$. Formally, we define the {\emph{maximal performance of $f$ in a neighborhood of $x$}} as:
\begin{equation*}
a_{\max,x} := \max_{U \in T_S,x \in U} a_U.
\end{equation*}
We define the {\emph{minimal performance of $f$ in a neighborhood of $x$}}, $a_{\min,x}$ analogously. As we show in Section \ref{sect-experiments}, these statistics can help illuminate the factors influencing the performance of a model on an individual test example.




\section{Experiments}
\label{sect-experiments}



\begin{table}
    \centering
    {\def\arraystretch{1.5}%
    \setlength{\tabcolsep}{4pt}
    \begin{tabular}{cc}
Open set & Accuracy\\
 \hline
\small{primary color: black $\cap$ rhinoceros auklet} & 39.13\\
\small{rhinoceros auklet} & 43.33\\
\small{bill length: same as head $\cap$ bill color: orange} & 46.03\\
\hline
\small{bill color: orange $\cap$ shape: duck-like} & 70.37\\
\small{throat color: black $\cap$ shape: duck-like} & 71.82\\
\small{bill shape: spatulate $\cap$ bill color: orange} & 72.72\\
\end{tabular}}
    \caption{A list of the $3$ neighborhoods of the image in Figure \ref{fig-bird-image} on which a ResNet18 model $f$ achieves its lowest accuracies (top) and the $3$ neighborhoods on which $f$ achieves its highest accuracies (bottom), all on the Caltech-UCSD Birds $200$ dataset.}
    \label{table-neighborhood-performance}
\end{table}

To explore some of the ideas described above, we apply our topological framework to the {\emph{Caltech-UCSD Birds $200$}} dataset \cite{WahCUB_200_2011}. This dataset has $11,788$ images of birds belonging to $200$ different species. We denote the training (respectively, test) set for this dataset by $D_{train}$ (resp. $D_{test}$). Critically for our task, this dataset also comes with $312$ binary attributes that, along with the bird classes themselves, can be used to generate a topology subbasis $S$. Specifically, $S$ consists of subsets $U_i$ where $i$ is some binary attribute for a bird in an image or a bird class. For example, is the eye color of the bird orange? Note that instances from the same class need not all have the same attributes. Even if a bird has orange eyes, if its eyes are not visible in an image, then it does not get included in $U_i$. As mentioned in Section \ref{sect-topology-on-dataset}, for this toy example we limit ourselves to at most one intersection of elements from the subbasis $S$ (so we only use elements of the form $U_i$ and $U_i\cap U_j$ for $U_i,U_j \in S$) and do not use intersections with less than $20$ elements, so in particular we only consider a small subset of the full topology, $\T_{birds}$. We train a ResNet18 convolutional neural network \cite{he2016deep} $f$ on $D_{train}$ starting from the Torchvision weights \cite{russakovsky2015imagenet} pretrained on ImageNet \cite{marcel2010torchvision}.

While across the entire Birds test set $f$ achieves an accuracy of  $0.571$ (i.e. $57.1\%$), our analysis reveals that $f$ has vastly different performance on different subpopulations. Surprisingly, out of the $11,798$ open sets we considered, $f$ achieved $1.000$ accuracy on $16$ of them and $0.000$ on $18$ of them. The majority of open sets with either the highest or lowest accuracy are the intersection of a class and an attribute. We note that among those open sets that are either a single attribute or an intersection of attributes, the bright colors of red, blue, and green seem to be associated with higher accuracy. For example ``underparts color: grey $\cap$ throat color: blue'' (accuracy $1.000$) or ``head pattern: capped $\cap$ nape color: red'' (accuracy $0.963$). 

We also analyze the local inconsistency of the set ``has throat color: yellow'', finding it to be quite high, at $0.546$ (with threshold $k$ set to $20$). This underscores the fact that the performance of $f$ can differ significantly when we restrict to open subsets of a subpopulation. For example, if $U$ corresponds to ``has throat color: yellow'' and $V$ corresponds to subset ``has breast color: yellow $\cap$ American goldfinch'' then $\res^\F_{U,V}(a_U)-a_V = -0.338$ while if $V$ corresponds to ``has belly pattern: solid $\cap$ common yellowthroat'' then $\res^\F_{U,V}(a_U)-a_V = 0.546$. We found greater inconsistency on subsets associated with attributes rather than classes, for example the local inconsistency for the class ``vermilion flycatcher'' was only $0.080$.

\begin{figure}[t]
\centering
\includegraphics[width=.6\columnwidth]{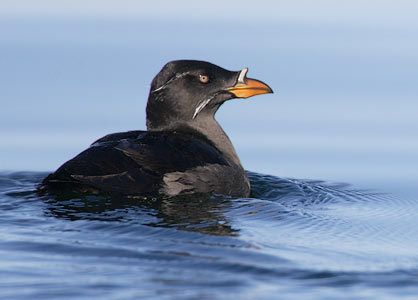} 
\caption{The rhinoceros auklet image on whose neighborhood we analyze the performance of ResNet18 model $f$ in Table \ref{table-neighborhood-performance}.}
\label{fig-bird-image}
\end{figure}

Finally, we demonstrate how the notion of the performance of a model in the neighborhood of a point can be used to better understand why $f$ handles a given input the way that it does. We choose the image, $x$, of the rhinoceros auklet shown in Figure \ref{fig-bird-image}. We see from Table \ref{table-neighborhood-performance} that the two neighborhoods where the model performs worst consists of those images that are characterized by being a rhinoceros auklet which does not have bright plumage. On the other hand, the model does well on neighborhoods characterized by either the shape of the bird or the bill shape/color (note though that the bill color is not exclusively found in top performing neighborhoods as indicated by the third worst performing open set containing $x$). 

For a more complete analysis, with more confident conclusions, a larger scale investigation including more of the topology $\T_{birds}$ would be required. We hope that the brief examples that we provide in this section help make the abstract constructions in this paper more clear.

\section{Conclusion}

In this paper we propose a new paradigm for model evaluation which is guided by the idea that a model's test set should be handled as a space rather than just a set. We show how this makes precise notions of ``local'' vs. ``global'' model performance, allowing a model trainer to better conceptualize the ways in which a model fails to be robust. We hope that this represents a first step toward bringing human's natural spatial intuition to bear on the challenge of evaluating complex machine learning models.

\bibliography{dataset_to_dataspace_arxiv}
\bibliographystyle{icml2021}

%
%
%

\end{document}